%% file: custom.tex
\pdfoutput=1

\documentclass[11pt]{article}

\usepackage[final]{acl}

\usepackage{times}
\usepackage{latexsym}
\usepackage{booktabs}
\usepackage{amsmath,amssymb}
\usepackage{tabularx} 
\usepackage{multirow}
\usepackage{graphicx}
\usepackage{arydshln}
\usepackage{colortbl}

\usepackage[tight,footnotesize]{subfigure}

\usepackage{enumitem}

\usepackage[T1]{fontenc}

\usepackage[utf8]{inputenc}

\usepackage{microtype}

\usepackage{inconsolata}

\usepackage{xspace}
\newcommand{\ourapproach}{\texttt{MR-MKG}\xspace}

\title{Multimodal Reasoning with Multimodal Knowledge Graph}

\author{
    Junlin Lee$^1$ \quad
    \quad Yequan Wang$^2$ \quad
    \quad Jing Li$^{1*}$ \quad
    \quad Min Zhang$^1$ \\
    $^{1}$Harbin Institute of Technology, Shenzhen, China \quad \\  $^{2}$Beijing Academy of Artificial Intelligence, Beijing, China \\
    \texttt{leejunlin27@gmail.com} \quad \texttt{tshwangyequan@gmail.com} \quad \\ \texttt{jingli.phd@hotmail.com} \quad \texttt{zhangmin2021@hit.edu.cn}
}

\begin{document}
\maketitle
\begin{abstract}
	

Multimodal reasoning with large language models (LLMs) often suffers from hallucinations and the presence of deficient or outdated knowledge within LLMs. 
Some approaches have sought to mitigate these issues by employing textual knowledge graphs, but their singular modality of knowledge limits comprehensive cross-modal understanding.
In this paper, we propose the Multimodal Reasoning with Multimodal Knowledge Graph (\ourapproach) method, which leverages multimodal knowledge graphs (MMKGs) to learn rich and semantic knowledge across modalities, significantly enhancing the multimodal reasoning capabilities of LLMs.
In particular, a relation graph attention network is utilized for encoding MMKGs and a cross-modal alignment module is designed for optimizing image-text alignment. 
A MMKG-grounded dataset is constructed to equip LLMs with initial expertise in multimodal reasoning through pretraining. 
Remarkably, \ourapproach achieves superior performance while training on only a small fraction of parameters, approximately 2.25\% of the LLM's parameter size.
Experimental results on \textit{multimodal question answering} and \textit{multimodal analogy reasoning} tasks demonstrate that our \ourapproach method outperforms previous state-of-the-art models.
\let\thefootnote\relax\footnotetext{$^*$ Corresponding author.}


%
%
%
%
%
%
%
%
%

\end{abstract}

\section{Introduction}

Recently, Large Language Models (LLMs)~\citep{gpt3, achiam2023gpt} have demonstrated their superiority and robustness across a variety of NLP tasks~\citep{zhang2023benchmarking, robinson2022leveraging,DBLP:journals/corr/abs-2310-00785}. 
To further unlock the potential of LLMs, researchers~\citep{abs-2303-04671, abs-2302-14045, abs-2205-02655, abs-2305-06355} have attempted to endow them with multimodal reasoning capabilities, as exemplified by visual LLMs like BLIP-2~\citep{0008LSH23}, MiniGPT-4~\citep{abs-2304-10592}, LLaVA~\citep{abs-2304-08485}, etc. 
Although these models have made significant strides in enabling reasoning with both images and text, they are still prone to \textit{hallucinations} ~\citep{Hallucination,DBLP:journals/corr/abs-2310-06827}, often caused by inadequate or outdated information.

\begin{figure}[t]
	\centering
	
	\includegraphics[width=1\columnwidth]{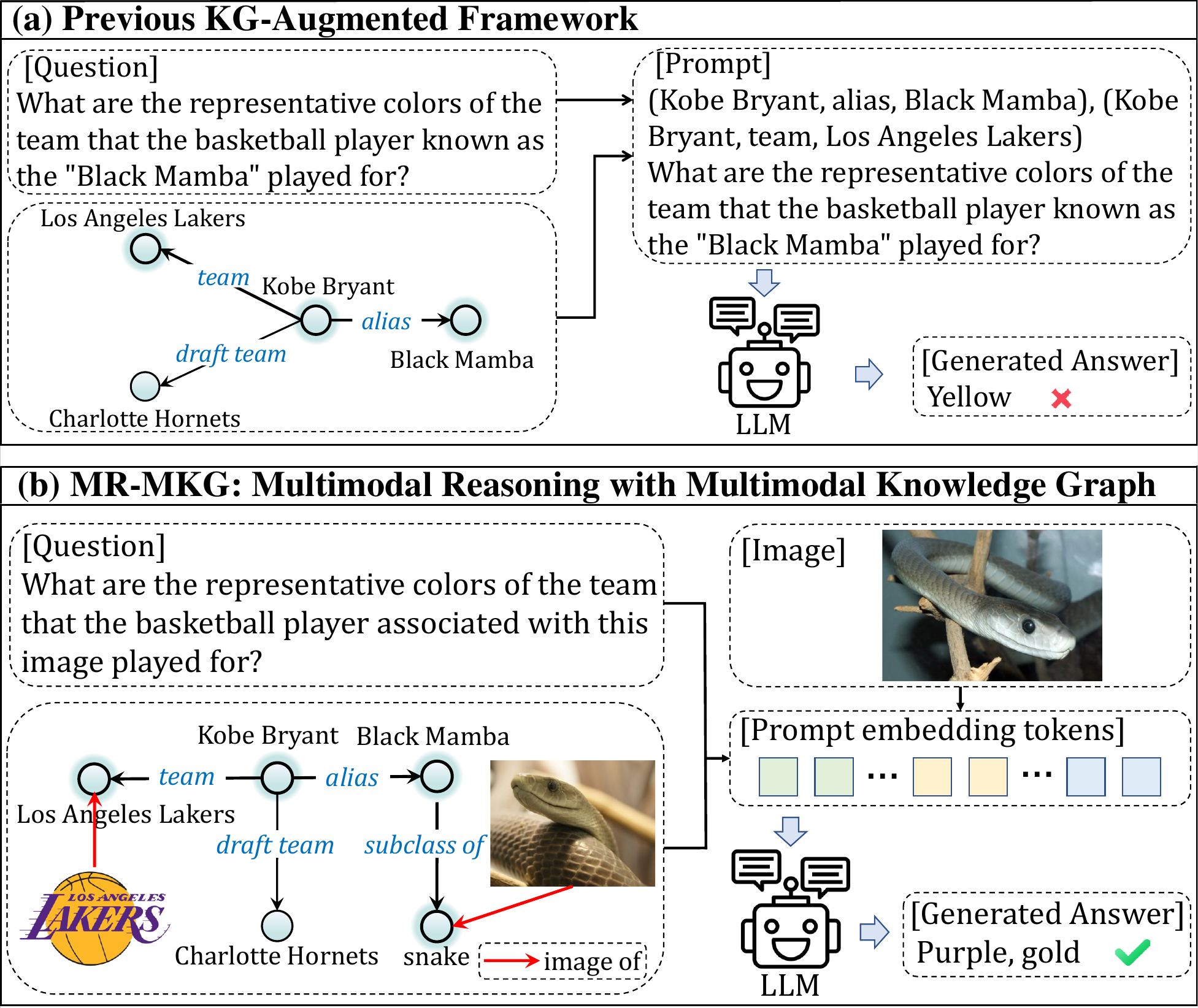}
	\caption{(a) The inadequate knowledge encapsulated within textual KG results in the incorrect answer. (b) Our \ourapproach produces the correct answer by reasoning with richer multimodal information. }
        \label{fig:motivation}
    \vspace{-4mm}
\end{figure}

Fine-tuning Large Language Models (LLMs) to update their knowledge base is often a time-consuming and costly process.
An alternative strategy, as suggested by~\citealt{wu2023retrieve}, involves leveraging knowledge graphs (KGs) as a means to directly augment LLMs with the requisite knowledge.
Although recent efforts~\cite{kaping,sen-etal-2023-knowledge,kim2023kg,DBLP:journals/corr/abs-2307-07697} have focused on employing textual KGs, their singular modality limits LLMs' ability to process and reason with multimodal information (as illustrated in Figure~\ref{fig:motivation}a). 
This limitation leads us to consider the use of multimodal knowledge graphs (MMKGs) instead of textual KGs (See Figure~\ref{fig:motivation}b).

In this paper, we propose the \textbf{M}ultimodal \textbf{R}easoning with \textbf{M}ultimodal \textbf{K}nowledge \textbf{G}raphs (\ourapproach) method, designed to expand the multimodal knowledge of LLMs by learning from MMKGs. 
In particular, \ourapproach first encodes the retrieved MMKG using a relation graph attention network (RGAT)~\cite{ishiwatari2020relation}, which generates knowledge node embeddings that are able to capture complex graph structures. 
Then, knowledge and visual adapter layers are designed to bridge the cross-modal gap, mapping both knowledge nodes and visual embeddings to word embedding of LLMs, respectively.
Finally, embeddings of knowledge nodes, image and text are concatenated to form the prompt and subsequently forwarded to LLMs to provide guidance and instructions. 
In addition, we introduce a novel cross-modal alignment module to optimize the image-text alignment through a matching task within MMKGs. 
To equip the model with initial expertise in multimodal reasoning, we first pretrain \ourapproach on a customized MMKG-grounded dataset, which is constructed by matching each VQA~\cite{krishna2017visual} instance with a corresponding MMKG, derived from the scene graph of its image and containing essential knowledge for answering questions.

To thoroughly evaluate our \ourapproach method, we conduct comprehensive experiments on  \textit{multimodal question answering}~\cite{lu2022learn} and \textit{multimodal analogy reasoning}~\cite{zhang2022multimodal}  tasks, spanning various LLM sizes and training configurations.
The experimental results confirm that \ourapproach effectively processes and utilizes knowledge from MMKGs for multimodal reasoning, outperforms previous state-of-the-art models with a 1.95\% increase in accuracy and a 10.4\% improvement in the Hits@1 metric. 
Importantly, \ourapproach freezes both LLM and the visual encoder, with only a small fraction of the parameters, approximately 2.25\% of the LLM's parameter size, being updated.
In summary, our main contributions are three-fold: 

\begin{itemize}[noitemsep,nolistsep]
	\item To the best of our knowledge, we are the first to investigate the problem of 
	expanding multimodal reasoning capabilities of LLMs by utilizing knowledge derived from MMKGs.
	\item We propose the \ourapproach method, specifically designed to extract valuable knowledge from MMKGs and seamlessly integrate multimodal information into LLMs. Additionally, we also develop a MMKG-grounded dataset for initially enhancing multimodal reasoning.
	\item We extensively evaluate \ourapproach on two multimodal reasoning tasks. \ourapproach achieves state-of-the-art performance by significant margins, outperforming recent baseline methods.
\end{itemize}





\section{Related Work}
\subsection{Multimodal Knowledge Graph}

The primary benefit of MMKG lies in their integration of additional modalities into traditional KGs.
By associating entities with related images or textual descriptions, MMKGs bring valuable visual and textual dimensions to the knowledge base, enhancing its ability to tackle complex tasks. 
For instance, approaches~\cite{IKRL, mousselly2018multimodal} integrate images with entity features in KGs, significantly improving entity representations for applications like knowledge graph completion and triple classification. \citealt{zhao2023boosting} introduce a method to enhance entity-aware image captioning through the use of MMKGs, where the MMKG associates visual objects with named entities and captures relationships between these entities. 
In the realm of recommendation systems, \citealt{sun2020multi} employ MMKGs, combining various data modalities such as images and texts, to enhance item representations.
Our approach differs from these existing solutions in that it stands as a pioneering effort in equipping LLMs with multimodal reasoning capabilities using MMKGs, rather than integrating MMKGs in a specific task.

\begin{figure*}[t]
	\centering
	\includegraphics[width=1\textwidth,draft=false]{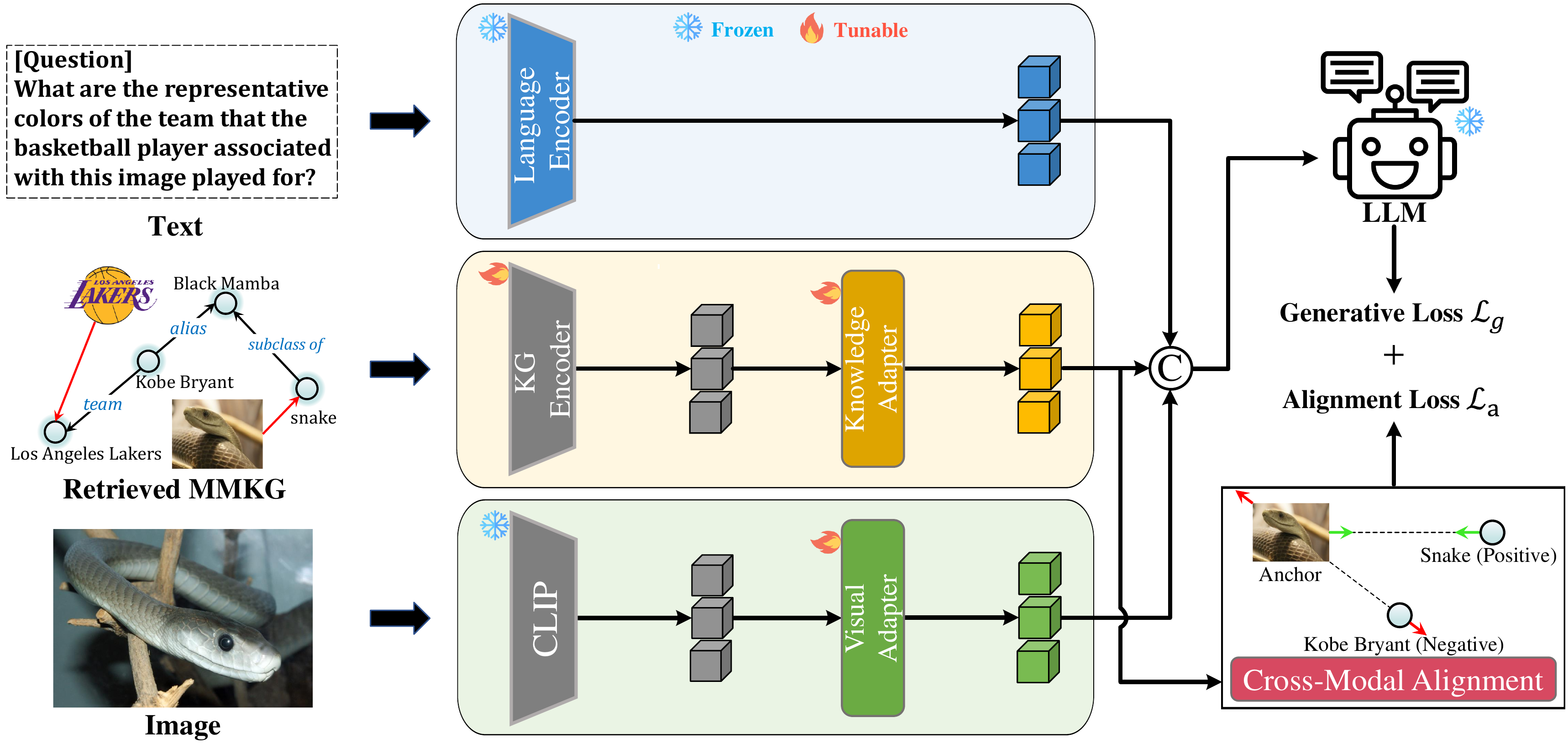}
	\caption{The overview of our \ourapproach approach. Text, multimodal knowledge graph, and image are independently embedded and then concatenated to form prompt embedding tokens. A cross-modal alignment module is designed to enhance the image-text alignment through a matching task within MMKGs.}
	\label{fig:model}
\end{figure*}

\subsection{Knowledge-Augmented LLMs}

While LLMs benefit from extensive pretraining on vast text corpora, they still face issues like hallucination and reliance on outdated knowledge, which hinder their reasoning abilities.
Consequently, recent studies~\cite{kaping, sen-etal-2023-knowledge, wu2023retrieve,mondal2024kam} have focused on incorporating knowledge directly into LLM prompts to mitigate these problems, thus eliminating the need for retraining the LLM. 
\citealt{kaping} extract relevant triples from KGs, converting them into text using linear verbalization techniques. 
\citealt{wu2023retrieve} develop a KG-to-Text approach for creating high-quality prompts, enhancing LLM performance in KG-based question answering by transforming relevant triples into more informative knowledge text.
\citealt{tian2023graph} observe that directly inputting triples from KGs into LLMs can introduce noise due to irrelevant contexts in KGs. They propose a graph neural prompt capable of extracting valuable knowledge from KGs for integration into pre-trained LLMs.
\citealt{mondal2024kam} incorporates external knowledge from text-based KGs into the multimodal chain of thought reasoning, enabling the model to achieve a deeper contextual understanding.
However, these methods primarily concentrate on textual KGs, which may limit their effectiveness in multimodal reasoning tasks due to the inherent differences in modalities.
To address this issue, we aim to enhance multimodal reasoning abilities by incorporating additional multimodal information from MMKGs.



 
 
 
 
 
 
  
  

\subsection{Multimodal Large Language Models}
The capabilities of purely text-based LLMs fall short of the evolving demands, leading to significant research efforts~\citep{abs-2303-04671, abs-2302-14045, abs-2205-02655, koh2023generating} aimed at developing LLMs proficient in handling multimodal inputs and tasks.
Current research trends~\cite{wu2023next, abs-2304-10592} primarily focus on integrating an adapter or projection layer to align the embedding spaces of various modal encoders with the text embedding space of the LLM. 
For example, popular visual LLMs like LLaVA~\citep{abs-2304-08485} and MiniGPT-4~\citep{abs-2304-10592} achieve this by freezing the LLM and training a visual projection to interpret visual data. 
This approach is mirrored in other multimodal LLMs, including auditory LLMs~\cite{zhang2023speechgpt} and video LLMs~\cite{zhang2023video}.
Recently, PandaGPT~\cite{pandagpt}, integrating the multimodal encoder ImageBind~\cite{girdhar2023imagebind}, is capable of understanding and processing six different modalities. Similarly, NExT-GPT~\cite{wu2023next} demonstrates proficiency in comprehending and generating content across four distinct modalities. 
However, these multimodal LLMs are still susceptible to hallucinations. While they enhance the alignment between modalities, they do not acquire new knowledge and may introduce new noise.
Our \ourapproach method differs from above methods in that the incorporation of MMKGs not only provides LLMs with additional, relevant information but also holds the promise of mitigating the noise generated during the transformation and alignment processes of multimodal data.

\section{Method}
In this section, we begin with an overview of  \ourapproach, followed by a detailed description on its architectural design and training approach.

\subsection{MR-MKG Overview}

The main objective of our method is to effectively leverage the capabilities of the Visual encoder and multimodal knowledge derived from MMKGs to enhance  the multimodal reasoning abilities of LLMs. 
A visual workflow is depicted in Figure~\ref{fig:model}.
Text, multimodal knowledge graph and image are independently embedded using a language encoder, KG encoder and visual encoder, respectively. 
The Visual and knowledge Adapters are designed to align the embedding spaces of visual and KG encoders with the text embedding space of the LLM.
The cross-modal alignment module is specifically designed to improve image-text alignment by utilizing a matching task within MMKGs.

\subsection{The MR-MKG Architecture}
\ourapproach consists of  five components: a language encoder, a visual encoder, a KG encoder, a knowledge adapter and a cross-modal alignment module. 
\paragraph{Language Encoder.}  
We adopt the embedding layers from  readily available LLMs like LLaMA and T5 as the language encoder, which remains fixed during both the training and inference phases.
Formally, the text is processed by the language encoder, resulting in text embedding $H_T$. 

\paragraph{Visual Encoder.} 
For an input image, we employ a pre-trained visual encoder like CLIP~\cite{radford2021learning}, which transfers the image into the visual feature ${X}_{I}$. 
To ensure compatibility between the visual and language space, a visual adapter implemented with a linear layer is used to transform the visual feature ${X_I}$ into visual-language embedding $H_I$, sharing the same dimensionality as the LLM's word embedding vector. 
Subsequently, a single-head attention network is utilized to obtain final visual features ${H}_{I}'$ associated with the text embedding $H_T$ by the following functions: 
\begin{equation}
	{H}_{I}  = {W}_{I} \cdot {X}_{I} +  {b}_{I}
\end{equation}
\begin{equation}
	{H}_{I}' =  \textrm{Softmax}(\frac{{H}_{T}{{H}_{I}^{\top}}}{\sqrt{d_k}}){H}_{I}
	\label{eq:selective_attn}
\end{equation}
where ${d}_{k}$ represents the dimension of ${H}_{T}$, and ${W}_{I}$ represents the trainable visual adapter matrix.

\paragraph{KG Encoder.} 
Given the text or image, \ourapproach first identifies related knowledge by retrieving a subgraph $\mathcal{G}$ from MMKG, which comprises the Top-$N$ most relevant triples.
However,  the retrieved subgraph $\mathcal{G}$  may also contain irrelevant triples, potentially introducing noise. 
Thus, if all these triples are directly fed into the prompt, the noise impedes the LLM's ability to efficiently process the essential knowledge. 
Additionally, the sequential prompt does not effectively capture the structural relationships in MMKG. 
Therefore, we employ the relation graph attention network (RGAT)~\cite{ishiwatari2020relation} to embed knowledge nodes by considering the intricate structures of $\mathcal{G}$. 
Specifically, we first use CLIP to initialize node and relation embeddings. 
Next, we use the RGAT network to encode $\mathcal{G}$ to generate knowledge node embeddings ${X}_{K}$. The process is as follows:
\begin{equation}
	{X}_{K} = f_{RGAT}(\mathcal{G})
\end{equation}

\paragraph{Knowledge Adapter.} 

To enable LLM to comprehend multimodal knowledge node embeddings, we introduce a knowledge adapter that transforms ${X}_{K}$ into text embeddings which are  understandable by LLMs. 
This knowledge adapter is designed to bridge the inherent gap between multimodal knowledge and text, fostering a more seamless alignment. 
Specifically, the node embeddings ${X}_{K}$ are mapped to knowledge-language embeddings ${H}_{K}'$ by:  
\begin{equation}
	{H}_{K}  = {W}_{K} \cdot {X}_{K} +  {b}_{K}
\end{equation}
\vspace{-0.25in}
\begin{equation}
	{H}_{K}' =  \textrm{Softmax}(\frac{\mathcal{Q}{{H}_{K}^{\top}}}{\sqrt{d_k}}){H}_{K}
	\label{eq:selective_attn}
\end{equation}
where ${W}_{K}$ represents the trainable knowledge adapter matrix, and $\mathcal{Q}$ corresponds to either ${H_T}$ or ${H_I}$, based on the specific scenario at hand.


\paragraph{Cross-Modal Alignment.} 
This module involves selecting a set of image entities from $\mathcal{G}$ at random and prompting the model to accurately match them with their corresponding textual entities. 
The knowledge node embeddings corresponding to the selected images are represented as ${H}'_{KI}$, and the embeddings for their associated text nodes are denoted as $H'_{KT}$.
We use the Triplet Loss~\cite{schroff2015facenet} for alignment. 
When the embeddings of one image entity $\{{H}'_{KI}\}_{i}$ serve as an anchor ${x}_{a}$,  its corresponding text entity embeddings $\{{H}'_{KT}\}_{i}$ serve as a positive sample ${x}_{p}$.
Concurrently, other text entity embeddings $\{{H}'_{KT}\}_{j\neq i}$ serve as negative samples ${x}_{n}$. 
The goal of alignment is to minimize the distance between positive samples and the anchor sample while maximizing the distance between negative samples and the anchor sample. 
The definition of alignment loss is as follows:
\begin{equation}
	\mathcal{L}_{a} = \sum_{i=1}^{M} \max(d({x}_{a}, {x}_{p}) - d({x}_{a}, {x}_{n}) + \alpha, 0)
\end{equation}
where $d$ represents the Euclidean distance, $M$ is the number of selected image entities, and $\alpha$ is a constant used to ensure a certain margin between the distances of positive and negative examples. 

\subsection{Training Objectives} 
The auto-regressive training objective focuses on training the LLM to predict subsequent tokens accurately. 
Specifically, we calculate the probability of generating the target answer $A$ by:
\begin{equation}
	\mathcal{L}_{g} =\sum_{i=1}^{L}\log p({A}_{i}|\textit{prompt},A_{0:i-1};\theta_a)
\end{equation}
where $L$ is the length of the target answer $A$, and $\textit{prompt}  = {H}_{K}' \oplus  {H}_{I}' \oplus  {H}_{T}$ is the concatenation of  visual embeddings ${H}_{I}'$, knowledge embeddings ${H}_{K}'$, and text embeddings ${H}_{T}$. $\theta_a$ denotes the adaptation parameters.The final objective function $\mathcal{L}$ is defined as the combination of $\mathcal{L}_{g}$ and $\mathcal{L}_{a}$:
\begin{equation}
	\mathcal{L} = \mathcal{L}_{g} + \lambda \mathcal{L}_{a}
\end{equation}
where  $\lambda$ is a trade-off weight for balancing two losses.
The training of \ourapproach is structured as a two-stage process. 
In the first stage, the model undergoes pre-training to develop foundational visual capabilities and to gain proficiency in understanding MMKGs. 
The second stage involves applying the model to specific scenarios that require advanced multimodal reasoning.
It is important to note that throughout both stages, the weights of both LLM and the visual encoder are unchanged.

\section{Experiments}
\subsection{Setups}

\paragraph{Evaluation Datasets.} We conduct experiments on multimodal question answering and multimodal analogy reasoning tasks, namely ScienceQA and MARS. See Appendix~\ref{app:data}  for additional details. 
\begin{itemize}[noitemsep,nolistsep]
\item \textbf{ScienceQA.}
This dataset is a large-scale multimodal science question answering dataset~\cite{lu2022learn}, each multiple-choice question is accompanied by a textual or visual context. This dataset is not purely multimodal, only 48.7\% of the data includes images.

\item \textbf{MARS.}
MARS~\cite{zhang2022multimodal} is a novel dataset designed for evaluating multimodal analogical reasoning over the multimodal knowledge graph MarKG. 
\end{itemize} 

\paragraph{Multimodal Knowledge Graphs.} 
In theory, any knowledge-rich MMKG can be applied to multiple benchmarks.
However, the suitable MMKG will vary depending on the benchmark's domain.
In particular, MMKG is used in conjunction with ScienceQA, whereas MarKG is employed to support MARS. The reason is provided in Appendix ~\ref{sec:mmkg}.

\begin{itemize}[noitemsep,nolistsep]
	\item \textbf{MMKG.}
This dataset~\cite{liu2019mmkg} is extracted from FreeBase~\cite{bordes2013translating}, DBpedia~\cite{auer2007dbpedia}, and YAGO~\cite{suchanek2007yago}, respectively. 
Each entity is associated with approximately 36 corresponding images from Google.
	\item \textbf{MarKG.}
MarKG~\cite{zhang2022multimodal} is a multimodal knowledge graph dataset developed from seed entities and relations in E-KAR~\cite{chen2022kar} and BATs~\cite{DBLP:conf/naacl/GladkovaDM16}. It aims to support MARS to do multimodal analogical reasoning,  sharing the same entity and relationship with MARS.
\end{itemize}

\paragraph{Pretraining setup.} 
We extract the image, QA pairs, and modified scene graph for each data instance to construct the MMKG-grounded dataset based on Visual Genome~\cite{krishna2017visual} for pretraining. 
Specifically, the object entities in the original scene graph are linked to their corresponding images and attributes through the ``\textit{image of}'' and ``\textit{attribute of}'' relations, respectively. More details are provided in Append~\ref{sec:dataset}.
This modified scene graph serves as the MMKG in pretraining.




\paragraph{Baselines.} 
For ScienceQA, we compare our approach against four kinds of baselines, including the zero- \& few-shot GPT Model~\cite{lu2022learn}, SOTA method MM-Cot~\cite{zhang2023multimodal}, representative end-to-end multimodal LLM model LLaVA~\citep{abs-2304-08485}, and parameter-efficient methods such as LLaMA-Adapter~\cite{zhang2023llama} and LaVIN~\cite{luo2023cheap}. 
For MARS, we compare our approach against two kinds of baselines, MKGE methods like IKRL~\cite{IKRL}, TransAE~\cite{TransAE}, and RSME~\cite{RSME}, Multimode pre-trained transformer model(MPT), VisualBERT~\cite{VisualBERT}, ViLT~\cite{ViLT}, and MKGformer~\cite{MKGformer}, etc. 
Each baseline undergoes pre-training on MarKG, equipping them with essential prior knowledge about entities and relations for enhancing multimodal reasoning.

\paragraph{Implementation.} 
We select the ViT-L/32~\cite{radford2021learning} as the visual encoder and RGAT as the knowledge embedding model for both datasets. 
In ScienceQA, we adopt FLAN-T5 3B and FLAN-T5 11B~\citep{chung2022scaling} as the LLMs and implement the  Multimodal-CoT prompting method~\cite{zhang2023multimodal}. 
To verify the generality of \ourapproach, FLAN-UL2 19B~\cite{chung2022scaling} is also used as the backbone. 
For MARS, LLaMA-2 7B~\cite{touvron2023llama} is selected to initialize our model. 
Regarding knowledge triple retrieval, we set the number of triples to either 10 or 20, and the hop distance for triple retrieval is maintained at one. 
All experiments are conducted on a NVIDIA 8$\times$A800-SXM4-80GB machine. 
More details are provided in Appendix ~\ref{sec:implementation}.

\begin{table*}[t!]
	\centering
	\resizebox{\textwidth}{!}{
		\begin{tabular}{lcccccccccc}
			\toprule
			
			\multirow{2}{*}{\textbf{Method}} &\multirow{2}{*}{\textbf{\#T-Param}} & \multicolumn{3}{c}{\textbf{Subject}} & \multicolumn{3}{c}{\textbf{Context Modality}} & \multicolumn{2}{c}{\textbf{Grade}} &  \multirow{2}{*}{\textbf{Average}} \\
			&& \textbf{NAT}   & \textbf{SOC}   & \textbf{LAN}   & \textbf{TXT}   & \textbf{IMG}   & \textbf{NO}    & \textbf{G1-6}  & \textbf{G7-12} &    \\ 
			\midrule
			
			Human~\cite{lu2022learn} &-&  90.23 & 84.97 & 87.48 & 89.60 & 87.50 & 88.10 & 91.59 & 82.42 & 88.40 \\
			GPT-3.5 (CoT)~\cite{lu2022learn} &-&  75.44 & 70.87 & 78.09 & 74.68 & 67.43 & 79.93 & 78.23 & 69.68 & 75.17 \\
			GPT-4~\cite{abs-2304-08485} &-& 84.06 & 73.45 & 87.36 & 81.87 & 70.75 & 90.73 & 84.69 & 79.10 & 82.69 \\ \midrule
			UnifiedQA$_{Base}$~\cite{lu2022learn} &223M& {71.00} &  {76.04} &  {78.91} &  {66.42} &  {66.53} & {81.81} &  {77.06} & 68.82 &  {74.11} \\
			UnifiedQA$_{Base}$(MM-CoT)~\cite{zhang2023multimodal} &223M&  87.52 & 77.17 & 85.82 & 87.88 & 82.90 & 86.83 & 84.65 & 85.37 & 84.91 \\
			UnifiedQA$_{Large}$(MM-CoT)~\cite{zhang2023multimodal} & 738M&  \underline{\textbf{95.91}} & 82.00 & \underline{90.82} & \underline{95.26} & \underline{88.80} & \underline{92.89} & \underline{92.44} & 90.31 & \underline{91.68} \\
			LLaVA~\cite{abs-2304-08485} &13B & 90.36 & \underline{\textbf{95.95}} & 88.00 & 89.49 & 88.00 & 90.66 & 90.93 & \underline{90.90} & 90.92 \\  \midrule
			LLaMA-Adapter~\cite{zhang2023llama} &1.8M & 84.37 &  88.30  &  84.36  & 83.72 &  80.32 &  86.90 &  85.83 &  84.05 & 85.19 \\
			LaVIN-7B~\cite{luo2023cheap} &3.8M & 89.25	&94.94 &	85.24&	88.51	&87.46	&88.08 &	90.16&	88.07 & 89.41\\
			LaVIN-13B~\cite{luo2023cheap} & 5.4M &  89.88&	94.49	&89.82	&88.95&	87.61&	91.85&	91.45&	89.72& 90.83\\
			\midrule
			
			\textbf{\ourapproach(FLAN-T5-3B)} &77M & 90.67 &  85.38  &  86.45  & 90.96 &  87.46 &  87.39 &  90.27 &  85.23 & 88.47 \\
			\textbf{\ourapproach(FLAN-T5-11B)} &248M &94.93	&90.1	&90.55	&94.53	&92.12	&92.2	&93.83	&90.9	&92.78\\
			\textbf{\ourapproach(FLAN-UL2-19B)} & 248M &  95.74 &	90.33	&\textbf{92.00}	&\textbf{95.50} &\textbf{92.41}&	\textbf{93.31}&	\textbf{93.98}&	\textbf{93.01}& \textbf{93.63}\\

			\bottomrule
			
		\end{tabular}
	}
	\caption{Results on the ScienceQA \textit{test} set with accuracy (\%). \#T-Params = number of trainable parameters. Question classes: NAT = natural science, SOC = social science, LAN = language science, TXT = text context, IMG = image context, NO = no context, G1-6 = grades 1-6, G7-12 = grades 7-12. Previous SOTA results are underlined. 
		The second segment: Zero- \& few-shot methods. The third segment: SOTA and representative models. The fourth segment: Parameter-efficient methods. The fifth segment: Our \ourapproach results. 
	}
		\label{tab:main}
\end{table*}

\subsection{Main Results}
\paragraph{Results on Multimodal Question Answering.}
Table~\ref{tab:main} reports the experimental results on ScienceQA. 
We can make the following observations:

First, our \ourapproach approach outperforms all baseline methods in terms of the average accuracy. 
The second segment of the table shows zero- and few-shot methods, even when applied to a popular LLM like GPT, still do not reach human-level performance. 
Notably, GPT-4, with its enhanced multimodal capabilities and larger parameter size, shows considerable improvements over GPT-3.5.
Although UnifiedQA$_{Large}$(MM-CoT) achieves previous SOTA, it requires training with its full parameters, leading to high training costs.
In contrast, \ourapproach requires training only a small fraction of the parameters and still achieve superior results. 
For instance, \ourapproach (FLAN-T5 3B) only trains 77M parameters but outperforms UnifiedQA\textsubscript{\textit{Base}} by 3.56\%, almost reaching human performance. 
\ourapproach (FLAN-T5 11B), with a comparable number of trainable parameters (249M) to UnifiedQA\textsubscript{\textit{Base}} (223M), achieves an absolute improvement of 7.87\% over it.
This indicates that our method achieves a more favorable balance between performance and training efficiency.

Second, although LLaVA achieves the best performance  in the SOC category, \ourapproach surpasses LLaVA in all other categories, with an average accuracy gain of +1.86\%.
Importantly, our \ourapproach method (FLAN-T5-11B), is trained on just 248M parameters, in contrast to LLaVA, which is trained on a much larger scale of 13 billion parameters.
We attribute this to the fact that \ourapproach is effective in enhancing multimodal reasoning, leveraging the multimodal knowledge derived from MMKGs. 


\begin{table}[t]
	\centering
	\begin{center}
		\resizebox{0.488\textwidth}{!}{
			\renewcommand{\arraystretch}{1.1}
			\begin{tabular}{lccccc}
				\toprule
				
				\textbf{Method}   & \textbf{Hits@1} & \textbf{Hits@3} & \textbf{Hits@5} & \textbf{Hits@10} & \textbf{MRR} \\
				\midrule
				
				IKRL~\cite{IKRL} & 0.266 & 0.294 & 0.301 & 0.310 & 0.283 \\
				TransAE~\cite{TransAE}   & 0.261 & 0.285 & 0.289 & 0.293 & 0.276 \\
				RSME~\cite{RSME} & 0.266 & 0.298 & 0.307 & 0.311 & 0.285  \\ \midrule
				
				MarT\_VisualBERT~\cite{VisualBERT}  & 0.261  & 0.292 & 0.308 & 0.321  & 0.284 \\
				MarT\_ViLT~\cite{ViLT} &0.245 & 0.275 & 0.287 & 0.303  & 0.266   \\
				MarT\_ViLBERT~\cite{ViLBERT} & 0.256  & 0.312 & 0.327  & 0.347  & 0.292 \\
				MarT\_FLAVA~\cite{FLAVA}  & 0.264 & 0.303 & 0.309 & 0.319 & 0.288  \\
				MarT\_MKGformer~\cite{MKGformer}  & \underline{0.301}  & \underline{0.367} & \underline{0.380} & \underline{0.408} & \underline{0.341} \\ \midrule
				
				\textbf{Visual\_LLaMA-2 7B} &0.286 &0.373 &0.409 & 0.457 &0.347   \\
				\textbf{\ourapproach (Visual\_LLaMA-2 7B)} &\textbf{0.405} &\textbf{0.465} &\textbf{0.497} &\textbf{0.531} &\textbf{0.449}    \\
				
				\bottomrule
			\end{tabular}
		}
	\end{center}
	\caption{\small  Results on the MARS \textit{test} set. The second segment: multimodal knowledge graph embedding (MKGE) methods. The third segment: multimodal pre-trained Transformer (MPT) methods. The fourth segment: \ourapproach. MarT indicates that models are pre-trained on MarKG. Visual\_LLaMA means that LLaMA is equipped with a visual adapter.}
	\label{tab:er}
\end{table}

Third, LLaMA-Adapter and LaVIN represent parameter-efficient approaches, focusing on training a lightweight adapter to assimilate different modal information.
In comparison, \ourapproach (FLAN-T5-11B) demonstrates significant superiority over these models, achieving absolute improvements of 7.59\% and 3.37\%, respectively.

\begin{table*}[t!]
	\small
	\centering
	\resizebox{\textwidth}{!}{
		\begin{tabular}{lccccccccc}
			\toprule
			
			\textbf{Settings}    & \textbf{NAT}   & \textbf{SOC}   & \textbf{LAN}   & \textbf{TXT}   & \textbf{IMG}   & \textbf{NO}    & \textbf{G1-6}  & \textbf{G7-12} & \multicolumn{1}{l}{\textbf{Average}} \\ \midrule
			
			Visual\_FLAN-T5-11B &88.45	&81.89	&84.09	&88.47	&86.51	&85.51	&86.75	&84.64	&86.08\tiny{{(+0.00)}}                  \\
			+ KG                            &93.78	&88.64	&89.55	&93.35	&90.47	&91.08	&92.77	&89.65	&91.74\tiny{(+5.66)}                  \\
			+ MMKG                            &94.23	&89.20	&90.00	&93.94	&91.77	&91.43	&93.39	&89.85	&92.21\tiny{(+6.13)}                  \\
			+ Alignment                            & 94.40	&89.54	&90.09	&94.18	&91.98	&91.50	&93.32	&90.38	&92.36\tiny{(+6.28)}                  \\
			+ Pre-training                            & 94.93	& 90.10	& 90.55	& 94.53	& 92.12	& 92.20	& 93.83	& 90.90	& 92.78\tiny{(+6.70)}                  \\
			
			\bottomrule
		\end{tabular}
	}
	\caption{Ablation study on the ScienceQA \textit{test} set. ``MMKG'' indicates using MMKG to replace KG. }
	\label{tab:ablation}
\end{table*}

\input{table/table8}

Fourth, to assess the generalizability of \ourapproach across different backbones, we experimented with LLMs of various sizes and types. 
The results in Table~\ref{tab:main}, reveal that expanding the parameters of the FLAN-T5 model from 3B to 13B leads to a significant performance boost, specifically an increase of +4.31\%. This suggests that larger models benefit more from our approach.
However, when both the backbone model and its parameters are altered, as with FLAN-UL2-19B, we observe the state-of-the-art performance, although the improvement margin is relatively modest. 
This could be attributed to the consistent number of training parameters or the inherent challenge in achieving higher accuracy improvements in already highly accurate models.


\paragraph{Results on Multimodal Analogical Reasoning.}
To further assess the generalizability of our \ourapproach method, we extend our experiments to a different task, i.e., multimodal analogical reasoning. 
The experimental results, as displayed in Table~\ref{tab:er}, clearly show that \ourapproach significantly outperforms all other methods on the MARS dataset. 
It is noteworthy that the performances of the multimodal knowledge graph embedding  methods and multimodal pre-trained Transformer models are relatively comparable, with MKGformer standing out with superior performance.
In contrast, the visual LLaMA-2 7B model, when equipped with a visual adapter, achieves results on par with MKGformer, albeit with a slightly lower Hits@1 score, but shows improvements in other metrics. 

This underscores the effectiveness and well-crafted design of the visual adapter component. 
Remarkably, when enhanced with \ourapproach, the visual LLaMA-2 7B exhibits a 10.4\% increase in terms of Hits@1 score, alongside significant improvements in other metrics.



 
  
  

\input{table/table9}
\subsection{Ablation Study}
To understand the impact of each component within \ourapproach method, we performed an ablation study on ScienceQA. 
As shown in Table~\ref{tab:ablation}, each component was added independently and their individual contributions were analyzed. 
The results in Table~\ref{tab:ablation} clearly illustrate the beneficial effect of each component on enhancing multimodal reasoning.

Most notably, the inclusion of knowledge extracted from KG results in the most substantial improvement, yielding a +5.66\% increase in performance. This highlights the pivotal role of KG-enhanced reasoning in the method. When multimodal knowledge from MMKG is incorporated, there is a further improvement in performance, rising from 91.74\% to 92.21\%. 
This indicates that multimodal knowledge effectively supplements the reasoning process with additional information.

The addition of an image-text matching task in the cross-modal alignment module  leads to a modest increase in accuracy to 92.36\%, underscoring its utility in refining the LLM's understanding of cross-modal information. Lastly, pre-training our model on the MMKG-grounded dataset brings an average accuracy improvement of +0.42\%, thereby demonstrating the advantages of pre-training.
In conclusion, these ablation studies distinctly validate the effectiveness of each component in the \ourapproach method, showing how they collectively contribute to the overall performance enhancement.

\begin{table}[t]
	\centering
	\small
	\begin{center}
		\resizebox{0.475\textwidth}{!}{
			\renewcommand{\arraystretch}{1.1}
			\begin{tabular}{lcc}
				
				\toprule
				
				\textbf{LLM}   & \textbf{Method} & \textbf{ScienceQA}  \\
				\midrule
				
				\multirow{3}{*}{\textbf{FLAN-T5-11B}\hspace{0.3cm}}
				& Text-Only & 92.78 \\
				& Image-Only & 91.58  \\
				& Text + Image & 92.03  \\
				
				\bottomrule
			\end{tabular}
		}
	\end{center}
        \caption{Average Accuracy(\%) with different subgraph retrieve methods on the ScienceQA \textit{test} set.}
	\label{tab:subgraph}
\end{table}

\begin{table}[t]
	\centering
	\begin{center}
		\resizebox{0.475\textwidth}{!}{
			\renewcommand{\arraystretch}{1}
			
			\begin{tabular}{lccc}
				\toprule
				
				\textbf{Model} &\textbf{Design}    & \textbf{ScienceQA}   & \textbf{MARS} \\ \midrule
				\multirow{3}{*}{\textbf{\begin{tabular}[c]{@{}c@{}}FLAN-T5-11B \\ /LLaMA-2 7B\end{tabular}}\hspace{0.3cm}}
				&GNN &92.23	&39.1 \\
				
				& GAT &91.94	&39.6 \\
				& RGAT &92.78	&40.5 \\
				
				\bottomrule
			\end{tabular}
		}
	\end{center}
	\caption{Impact of different KGE architectures. The metric is Average Accuracy and Hits@1, respectively.}
	\label{tab:gnn}
\end{table}

\begin{figure}[t]
	\centering
	\subfigure{ \includegraphics[width=0.23\textwidth,draft=false]{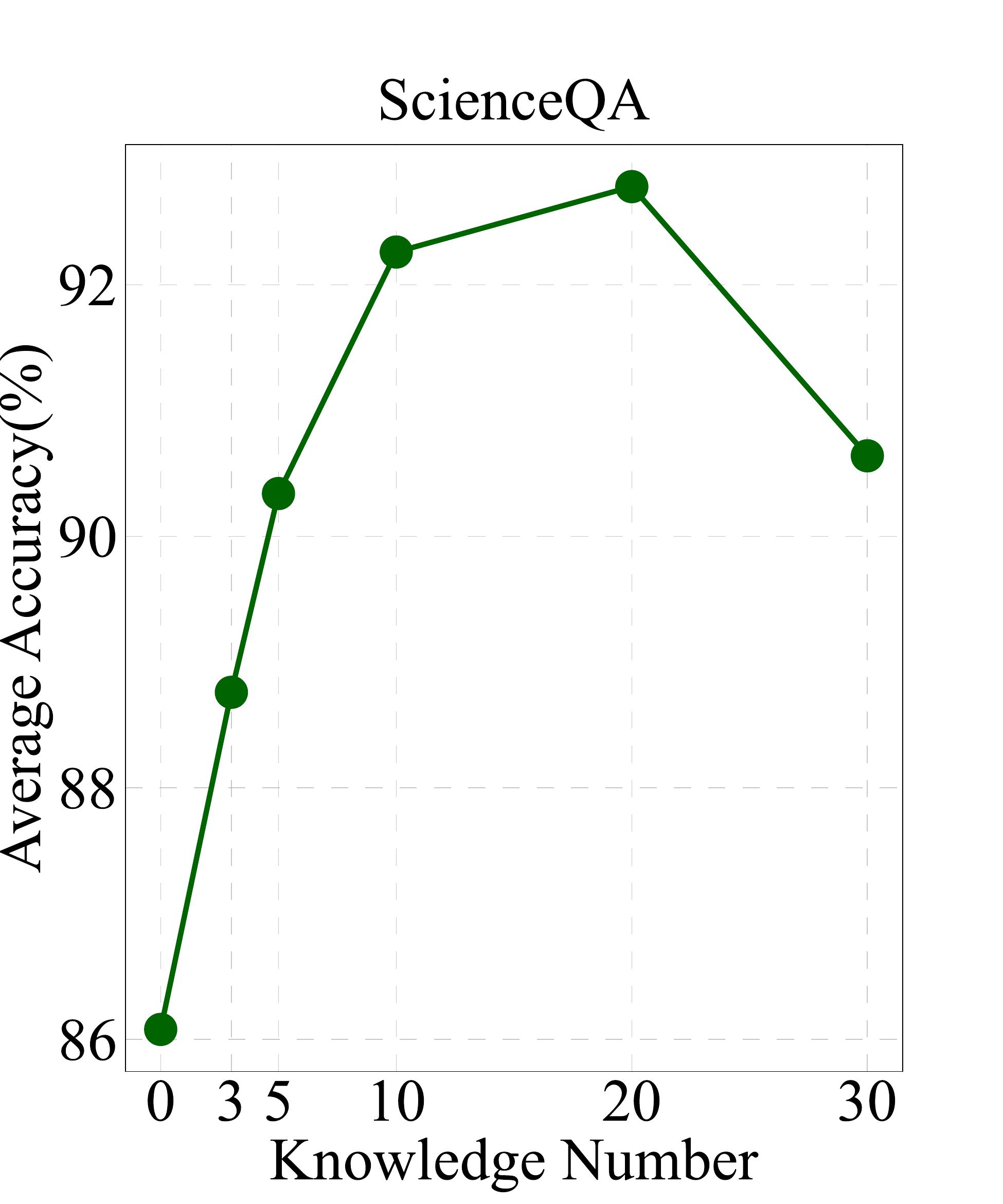}}  
	\subfigure	{\includegraphics[width=0.23\textwidth, draft=false]{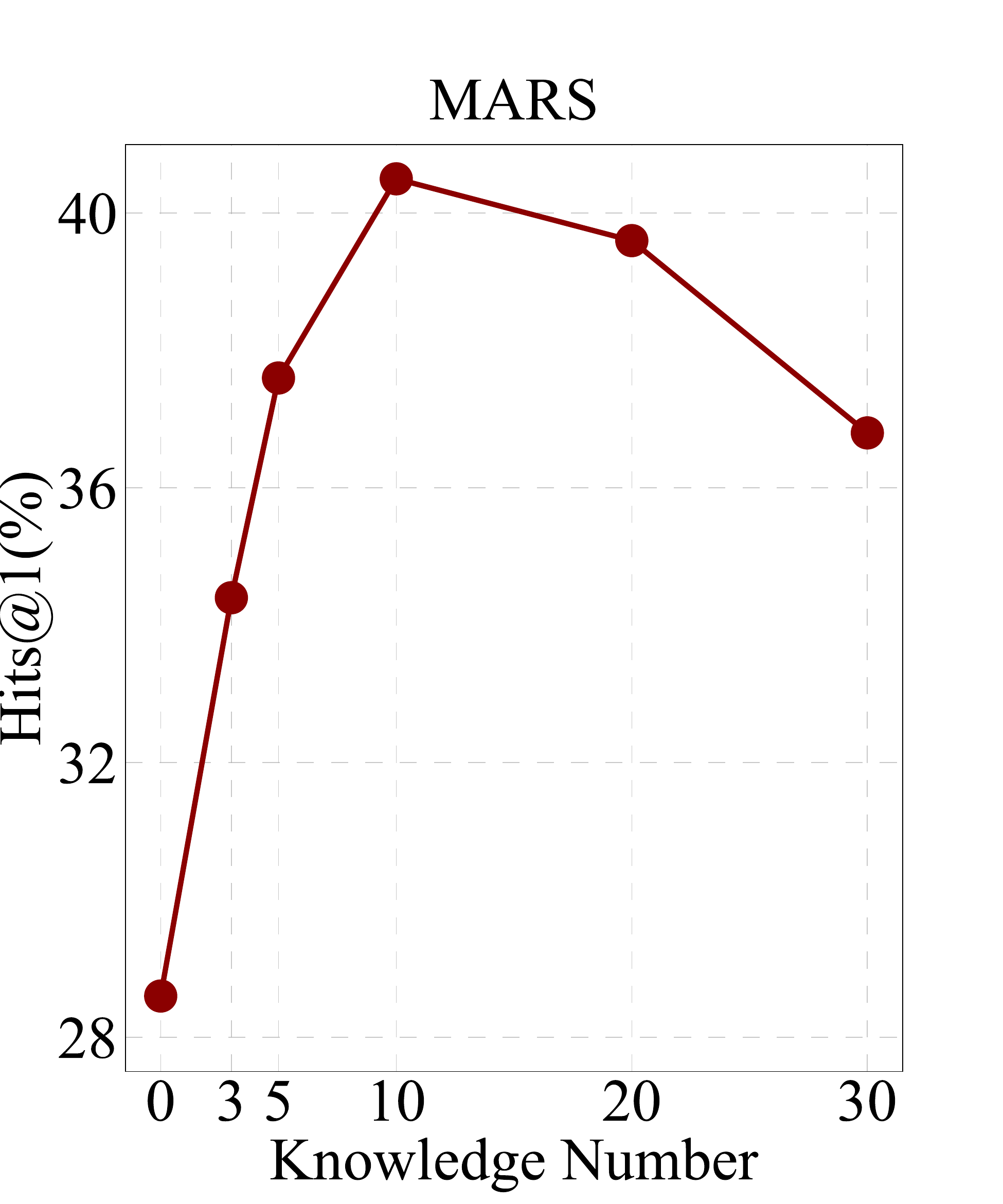} } 
	\caption{Impact of numbers of knowledge triplets. }
	\label{fig:kgnumber}
\end{figure}

However, we observe that the impact of MMKG and cross-modal alignment is relatively marginal. This is because ScienceQA is primarily text-oriented. As a QA dataset, its core questions and choices are presented in text form, resulting in fewer questions that require visual knowledge to answer. Additionally, ScienceQA is not entirely multimodal, with only 48.7\% of the data containing images, which further diminishes the true effectiveness of MMKG and cross-modal alignment. Moreover, when the model achieves higher accuracy, further increasing accuracy becomes challenging, resulting in less pronounced changes.

To prove the true effectiveness of MMKG and cross-modal alignment, we manually selected 1973 samples from ScienceQA. These samples all contain images, and their subjects are social science or natural science. We hypothesize that these samples require visual knowledge to reason the answer.

Table~\ref{tab:samples} shows the additional ablation study results on these samples. We can observe that the utilization of KG yields a 3.78\% increase in performance, and the use of MMKG yields a 1.41\% improvement. The addition of cross-modal alignment results in a performance improvement of 0.54\%. Compared to the improvements of using MMKG (0.47\%) and cross-modal alignment (0.15\%) in the original ablation study, the performance gains (1.41\% \& 0.54\%) from these samples are more significant. This confirms the true effectiveness of MMKG and cross-modal alignment.

In addition, We also supplement the results of the ablation study on MARS. Table~\ref{tab:MARS} shows that when using MMKG and cross-modal alignment, the model's performance significantly improved (2.9\% \& 1.3\%). These improvements relative to the improvement from using KG (6.6\%) are also noticeable. Therefore, the performance gain is relatively significant when the model's performance is low. Conversely, improving performance becomes much more challenging when the model's performance is already high.
\begin{figure}[t]
	\centering
	\subfigure{ \includegraphics[width=0.23\textwidth,draft=false]{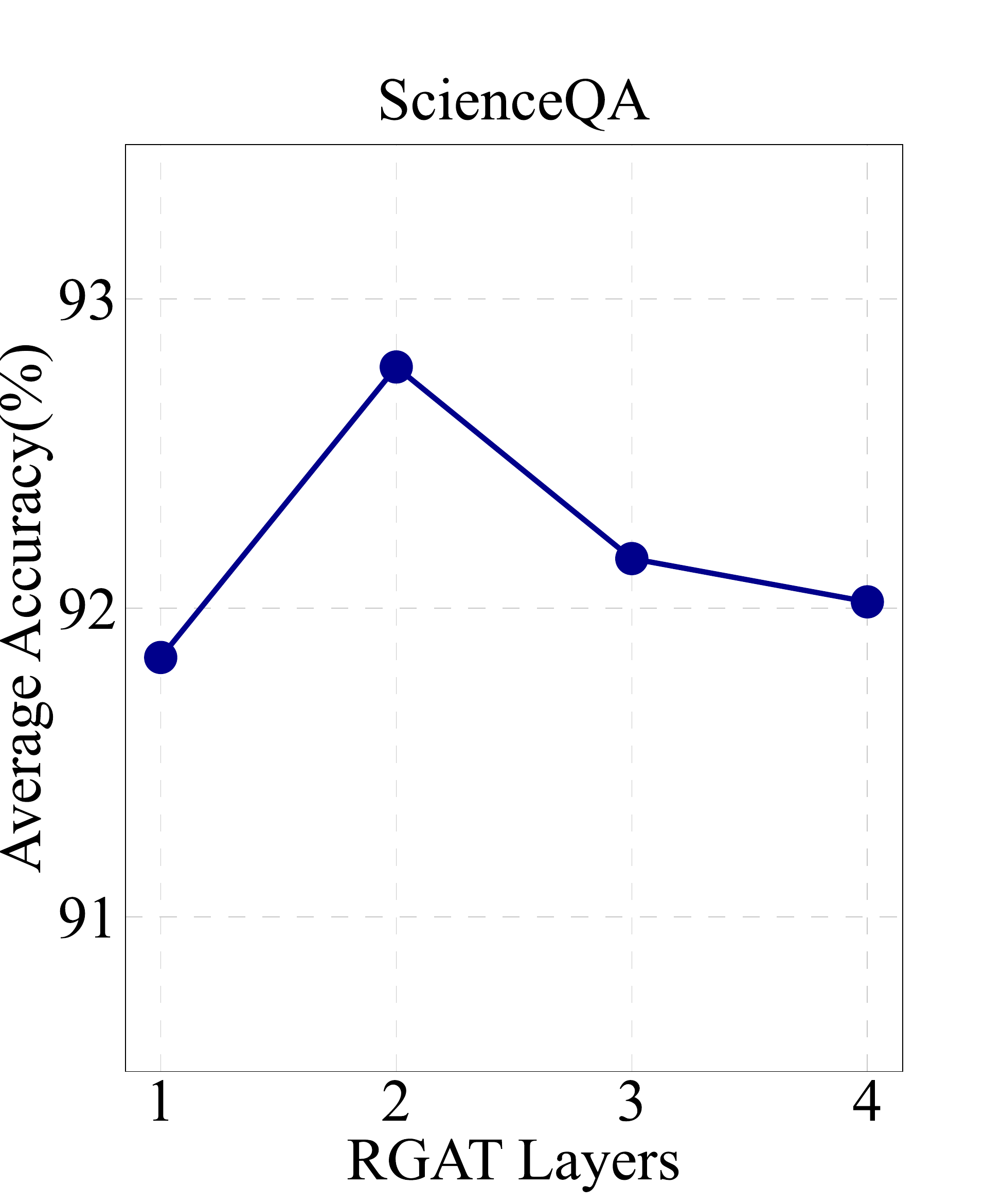}}  
	\subfigure	{\includegraphics[width=0.23\textwidth, draft=false]{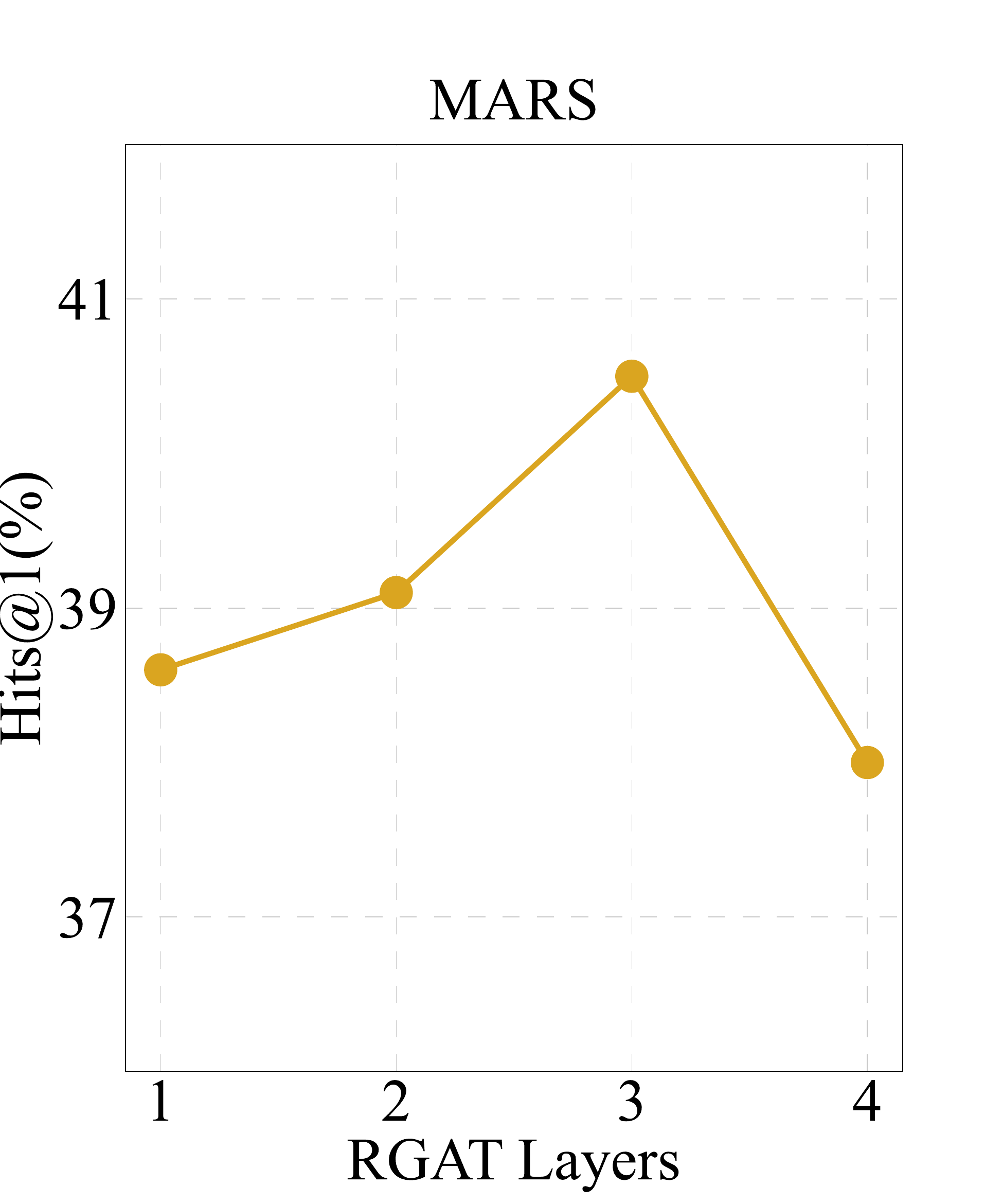} } 
	\caption{Impact of numbers of KGE layers. }
	\label{fig:layers}
\end{figure}

\subsection{Further Analysis}
In this section, we quantitatively investigate the impact of various architectural choices. 

\begin{figure*}[t]
	\centering
	\includegraphics[width=1\textwidth,draft=false]{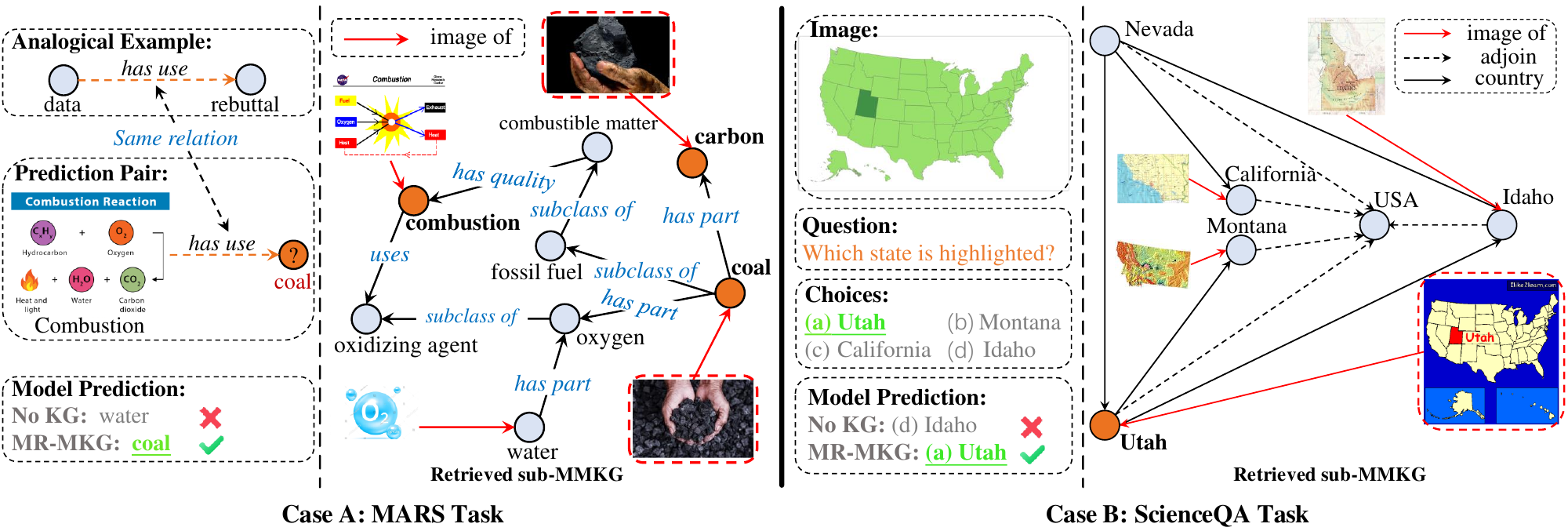}
	\caption{Two examples from MRAS and scienceQA datasets. In case A, the model needs to predict \textbf{coal} based on an Analogical Example and the image of combustion. In case B, the model needs to select the correct answer based on the image and the question. Relevant entities for reasoning are marked in orange or highlighted with a red box.}
		\label{fig:case}
\end{figure*}

\paragraph{Impact of  different subgraph retrieval methods.}
The experimental results, as detailed in Table~\ref{tab:subgraph}, indicate that the text-only retrieval strategy is the most effective, followed by the combined text and image strategy, while the image-only approach yields the least favorable results.
This pattern can be attributed to the characteristics of the ScienceQA dataset. 
This finding underscores the importance of tailoring the retrieval strategy to the specific nature of the problem at hand, rather than relying exclusively on one particular modality.

\paragraph{Impact of different knowledge graph embedding methods.}
We experimented with three distinct KGE architectures:  GNN~\cite{scarselli2008graph}, GAT~\cite{velivckovic2017graph}, and RGAT~\cite{ishiwatari2020relation}.
As shown in Table~\ref{tab:gnn}, the performances of GNN and GAT are quite comparable across both tasks, albeit slightly trailing behind RGAT. 
Notably, RGAT demonstrates the best performance in both tasks, underlining its efficacy as a widely adopted GNN architecture for explicitly modeling relationships in graph data.

\paragraph{Impact of different numbers of knowledge triplets.}
As depicted in Figure~\ref{fig:kgnumber}, we observe that as the number of triplets increases from 0 to 10, there is a proportional improvement in the performance of both models. 
However, an interesting trend emerges as the number of triplets extends beyond 20 to 30. In this range, we notice a decline in performance for both models. 
This decline implies that the quantity of useful knowledge triplets within the MMKG is limited, and an excess of triplets can introduce irrelevant information.

\paragraph{Impact of different numbers of KGE layers.}
Figure~\ref{fig:layers} demonstrates our exploration into the impact of varying the number of layers in RGAT. 
The trend indicates that an appropriate stacking of RGAT layers can positively affect the encoding of graph structures and representation of knowledge.

\subsection{Qualitative Analysis}
Figure~\ref{fig:case} (and Figure~\ref{fig:Example of ScienceQA} \& \ref{fig:Example of MARS} in Appendix) visualize the retrieved sub-MMKG for each task. For visual clarity, we only show relevant entities and relations. 
In MARS, the model aims to predict ``coal'' based on the image of combustion and an example of (data, rebuttal). 
Our \ourapproach approach identifies and retrieves entities like ``combustion'', ``carbon'' ``water'', and ``oxygen'' from the image. 
The sub-MMKG provides an indirect connection linking ``combustion'' with ``coal''. 
The similarity between carbon and coal images guides the model to the correct prediction of  ``coal'', demonstrating the pivotal role of multimodal knowledge from MMKGs.

In the ScienceQA example, where the question is ``\textit{Which state is highlighted?}'', the model must identify this state's shape. 
Lacking sufficient intrinsic knowledge, the model without KG inaccurately predicts ``Idaho''. 
However, the sub-MMKG retrieved under \ourapproach holds crucial information about the shapes of different states in the options, directly informing the model about Utah's shape.
Both of these examples demonstrate the effectiveness of the multimodal knowledge derived from MMKGs.

\section{Conclusion}
In this study, we addressed the challenge of enhancing the multimodal reasoning capabilities of LLMs through the use of multimodal knowledge graphs. 
Our proposed approach, termed \ourapproach, is designed to empower LLMs with advanced multimodal reasoning skills by harnessing the rich knowledge (image, text and knowledge triplets) contained in MMKGs.
Comprehensive experiments on \textit{multimodal question answering} and \textit{multimodal analogy reasoning} tasks demonstrated the effectiveness of our \ourapproach approach, achieving the new state-of-the-art results in these tasks.
Furthermore, we also conducted a series of ablation studies, analytical examinations, and case studies to provide additional evidences of effectiveness.

\section*{Acknowledgements}
This work was supported in part by NSFC (U23B2055), Shenzhen College Stability Support Plan (GXWD20231128103232001), Department of Science and Technology of Guangdong (2024A1515011540) , National Science and Technology Major Project (2022ZD0116314) and NSFC (62106249).

\section*{Limitations}
In this section, we faithfully discuss the limitations that we would like to improve in future work. 


First, the efficacy of the retrieved sub-multimodal knowledge graph is contingent upon the success of the knowledge retrieval strategy employed. Should the retrieval scheme prove ineffective or underperform, it risks failing to procure pertinent knowledge for the posed question. This shortfall directly diminishes the probability of the LLM yielding accurate responses (refer to Figure~\ref{fig:Error example} in the Appendix and Table~\ref{tab:subgraph} for instances of such errors). Thus, a pivotal direction for future research involves refining the retrieval scheme to ensure it can supply the necessary and more precise knowledge essential for multimodal reasoning tasks.


Second, due to constraints in computational resources, our evaluation was limited to four LLMs across two multimodal reasoning tasks--ScienceQA and MARS.
However, there are still many LLMs with larger parameter sizes, such as LLaMA-2 70B~\cite{touvron2023llama}. Therefore, one of the future works is to scale up our method to even larger model sizes and assess its performance on a broader range of multimodal reasoning tasks.


\section*{Ethical Considerations}
Due to the limited knowledge retrieval capabilities and the potential for errors or outdated knowledge, the performance of our \ourapproach method is not yet perfect. Our approach has been evaluated on two publicly available datasets, ScienceQA and MARS. We explicitly claim that the applicability of our method and findings may be confined to similar datasets or domains. The performance of our method on other specific datasets or domains remains uncertain. Thus, there are potential risks when applying our method to privacy-sensitive or high-risk datasets. We should be cautious and verify whether the method generates correct answers.

\bibliography{custom}

\newpage

\appendix



\section{Additional Experimental Setups}
\label{sec:appendix}
\subsection{Datasets}
\label{app:data}
We provide additional details for two 
multimodal reasoning datasets, namely ScienceQA and MARS.
\paragraph{ScienceQA.}
ScienceQA is divided into training, validation, and test sets, consisting of 12,726, 4,241, and 4,241 instances, respectively. This dataset includes rich annotations, with lectures and explanations provided for the answers and context provided for the question. This dataset is unique in its combination of multimodal questions (text and image contexts) and its extensive coverage of 26 topics, 127 categories, and 379 skills. 
\paragraph{MARS.}
MARS features a mix of visual and textual modalities and is underpinned by the multimodal knowledge graph MarKG. MARS is among 2063 entities and 27 relations, each of these entities has corresponding images. It includes 10,685 training, 1,228 validation, and 1,415 test instances. 

\subsection{Multimodal Knowledge Graphs}
\label{sec:mmkg}
\input{table/table7}
\paragraph{MMKG.}
MMKG~\cite{liu2019mmkg} is constructed using FB15K as a template to generate the multimodal knowledge graph. Alignment of entities from FB15K with those in other knowledge graphs is achieved through the use of \textit{sameAs} links, resulting in DB15K and YAGO15K. In total, there are 812,899 triples among 14,951 entities and 1,345 relations. 
Most entities have corresponding images.
\paragraph{MarKG.}
MarKG~\cite{zhang2022multimodal} comprises 11,292 entities, 192 relations, and 76,424 images. The image data is sourced through Google searches and queries from the multimodal data Laion-5B~\cite{schuhmann2022laion}, using text descriptions of entities.

MarKG is chosen to support MARS to do multimodal analogical reasoning because MarKG and MARS come from the same work~\cite{zhang2022multimodal}. They have identical data sources and construction methods, sharing the same entities and relationships. Therefore, we hypothesize that the knowledge contained in MarKG is highly compatible with MARS and is more suitable for MARS compared to MMKG~\cite{liu2019mmkg}. Additionally, we replace MarKG with MMKG to validate this hypothesis.

Table~\ref{tab:MMKGs} shows that when MarKG is replaced with MMKG, the model's performance decreases from 40.5\% to 38.4\%. This indicates that the knowledge contained in MarKG is more crucial for MARS, but it doesn't imply that MMKG is ineffective, it still significantly improves the model's performance. So any knowledge-rich MMKG can be applied to multiple benchmarks. However, their enhancement for specific tasks may not be as effective as specific MMKG, but they can still improve the model to some extent. Therefore, the only selection criterion for MMKG is whether the knowledge required for the task is sufficient. For ScienceQA, we directly use MMKG to support it.
\input{figures/vg}
\subsection{MMKG-Grounded Dataset Construction}
\label{sec:dataset}
Visual Genome~\cite{krishna2017visual} is a large-scale image semantic understanding dataset. It consists of five main components for each data: Question-Answer pair, Region Description, Region Graph, Scene Graph, and Image. Each image is segmented into multiple regions, each of which is described separately. All objects and relationships within the image are extracted to construct the Scene graph. Two types of QA pairs are annotated: 1. \textbf{Freeform QA} based on the entire image (without specifying a region), and 2. \textbf{Region-based QA} based on the selected regions within the image.

We construct the MMKG-grounded dataset by extracting the image, QA pairs, and modified scene graph for each data instance. In this context, the combination of the image and the associated questions and answers constitutes a Visual Question Answering (VQA) task, representing multimodal reasoning. The modified scene graph functions as a multimodal knowledge graph, encompassing knowledge about the objects in the image (as shown in Figure~\ref{fig:vg}). Specifically, we crop out images of objects based on their bounding boxes and link them to their corresponding object entities using the ``\textit{image of}'' relation. Additionally, connections are established between object attributes and their respective entities through the ``\textit{attribute of}'' relation. Finally, we exclusively opt for Region-based QA data, as the corresponding multimodal scene graph contains key knowledge for reasoning out the answers. In total, we constructed 18,448 instances.

\input{table/table6}

\subsection{Large Language Models}
We describe the specific details of Large Language Models (LLMs) that we used for evaluation.
\paragraph{FLAN-T5.}
T5~\cite{raffel2020exploring} is an encoder-decoder model. For the same number of parameters, FLAN-T5~\cite{chung2022scaling} has been fine-tuned based on T5 on more than 1000 additional tasks covering a wider range of languages. 

\paragraph{FLAN-UL2.}
FLAN-UL2~\cite{chung2022scaling} expands on the FLAN-T5, using the upgrade pre-training process of UL2~\cite{tay2022ul2} which is a unified framework for pretraining models. 

\paragraph{LLaMA-2.}
LLaMA-2~\cite{touvron2023llama} is a free and open-source decoder-only model.

\begin{figure*}[t]
	\centering
	\includegraphics[width=0.95\textwidth,draft=false]{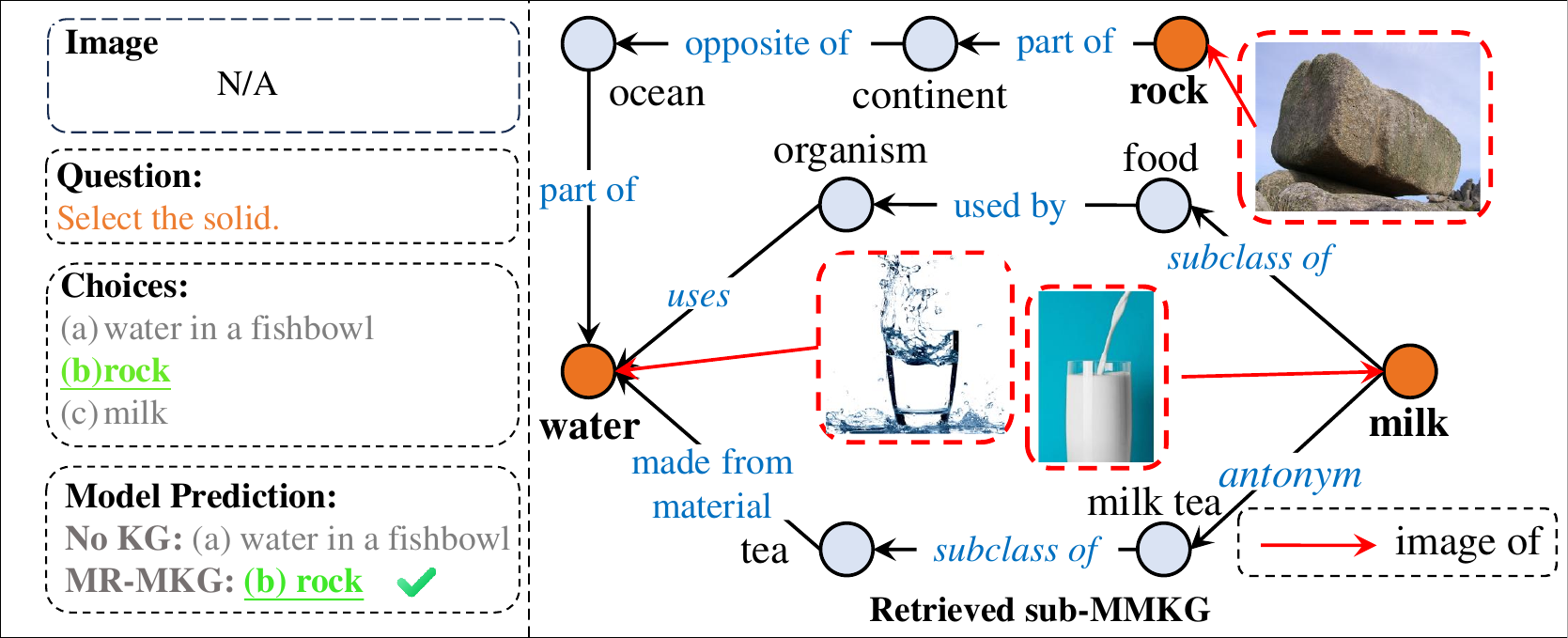}
	\caption{Example of ScienceQA.}
		\label{fig:Example of ScienceQA}

\end{figure*}

\subsection{Detailed Evaluation Metrics}
For the ScienceQA dataset, we only use accuracy as the evaluation metric. For the MARS dataset, we use Hits@k and MRR as our evaluation metrics. These metrics are all within the [0,1] range. A higher value indicates better performance. The Hits@k metric is obtained by calculating the number of times the correct entity appears at the first k positions in the predictions.  Denote the rank of the correct entity of $i$ triple as ${rank}_{i}$, and the reciprocal rank is 1/\textit{${rank}_{i}$}. The Mean Reciprocal Rank (MRR) is the average of the reciprocal ranks across all triples in the multimodal knowledge graph:

\begin{equation}
    \text{MRR} = \frac{1}{|\mathcal{M}|} \sum\limits_{i}^{|\mathcal{M}|}\frac{1}{rank_i}
\end{equation}
where $|\mathcal{M}|$ is the total number of the training set.

\subsection{Knowledge Retrieve Schemes}
The sub-MMKG $\mathcal{G}$ is retrieved based on the text or image information. This involves embedding the text or image information along with all the triples from the MMKG into the representation space. The cosine similarity between them is then computed, and all the entities of the Top-$n$ relevant triples form $E'$. Subsequently, $\mathcal{G}$ is retrieved based on the entities in $E'$, encompassing their one-hop neighbors and the relations connecting them~\cite{dragon}. Finally, we select the Top-$N$ most relevant triples in $\mathcal{G}$ based on cosine similarity.

\begin{figure*}[t]
	\centering
	\includegraphics[width=0.95\textwidth,draft=false]{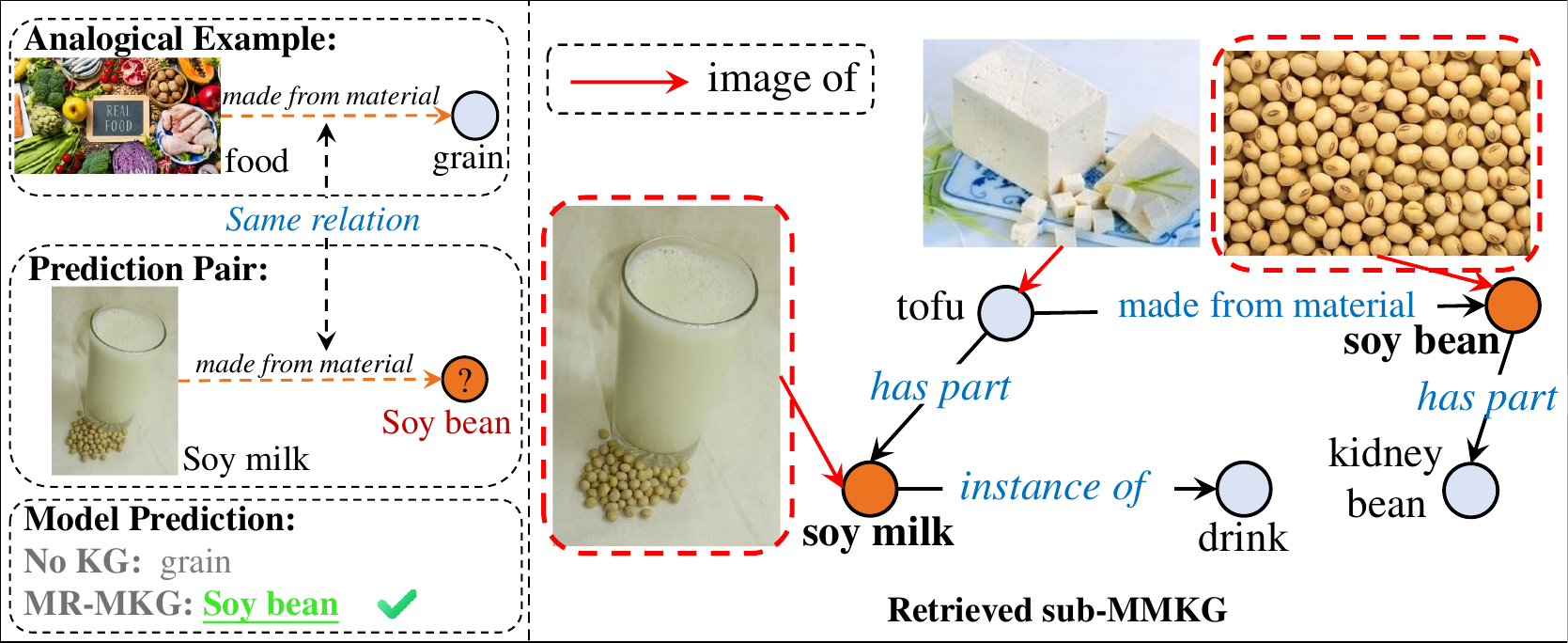}
	\caption{Example of MARS.}
		\label{fig:Example of MARS}
\end{figure*}

\subsection{Implementation Details}
\label{sec:implementation}
\paragraph{ScienceQA.}
We use Multimodal-CoT prompting in ScienceQA task. It is a two-stage framework that separates rationale generation and answer inference. In each prediction, the model initially generates a rationale and then predicts the final answer based on the question and the rationale. 
Relevant sub-MMKGs are retrieved based on text features, which are concatenated by question, context, and options.
We fine-tune the models up to 3 epochs, with a learning rate of 4e-5. We set the maximum number of input and output token lengths of LLMs is 512. The batch size is 1 and the optimizer is AdamW with a weight decay of 0.01.

\begin{figure*}[t]
	\centering
	\includegraphics[width=0.9\textwidth,draft=false]{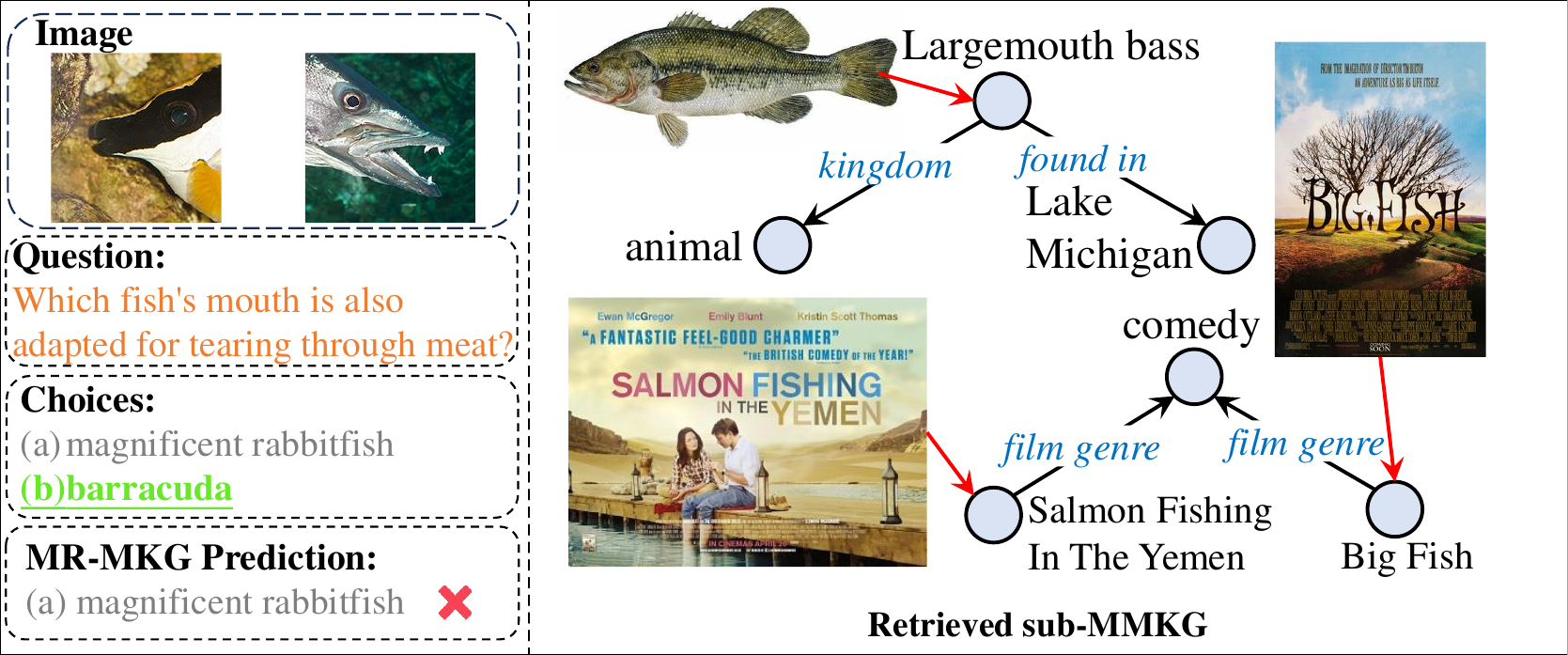}
	\caption{Error example.}
		\label{fig:Error example}
\end{figure*}

\paragraph{MARS.}
Initially, we pre-train the models on the MarKG dataset, focusing on tasks such as entity prediction and relation prediction. It can acquire entity and relation embeddings matrix. Subsequently, we further fine-tune the models on MARS. The details of hyper-parameters can be seen in Table~\ref{tab:parameter}. Relevant sub-MMKGs are retrieved based on the question entity, and all entities are embedded and transformed according to their modes.

\paragraph{MMKG-Grounded Dataset.}
We first pre-train our \ourapproach method on the MMKG-grounded dataset up to 2 epochs, with a learning rate of 5e-5. We set the maximum number of input and output token lengths of LLMs are 512 and 128, respectively. The batch size is 2 and the optimizer is AdamW with a weight decay of 0.01. Relevant sub-MMKGs are retrieved based on the question.

\section{Additional Examples of Case Studies}
To better understand the behavior of \ourapproach, we provide additional examples for case analysis.
\paragraph{Additional Case Studies.}
Figure~\ref{fig:Example of ScienceQA} and Figure~\ref{fig:Example of MARS} are additional examples. In Figure~\ref{fig:Example of ScienceQA}, the model needs to choose the solid among the options: ``rock'', ``milk'', and ``water in a fishbowl''. \ourapproach retrieves these entities and constructs the sub-MMKG. This sub-MMKG directly supplies images of these entities, enabling the model to better distinguish which option is the solid and obtain the correct answer, unaffected by ``fishbowl'' in option a. In Figure~\ref{fig:Example of MARS}, the model aims to predict ``soy bean'' based on the image of soy milk and an example of (image of food, grain). The sub-MMKG provides an indirect connection linking ``soy bean'' with ``soy milk'' through the intermediary ``tofu''. Within the sub-MMKG, the image of soy milk directly incorporates visual features of soy bean that closely resemble the image of soy bean itself. This guides the model to accurately predict ``soy bean'', highlighting the pivotal role of multimodal knowledge from MMKGs in multimodal reasoning.

\paragraph{Error Analysis.}
we also conduct an error case study, as illustrated in Figure~\ref{fig:Error example}. In this instance, the question is ``Which fish's mouth is also adapted for tearing through meat?''. However, the sub-MMKG retrieved by \ourapproach does not contain any useful information. Specifically, it contains knowledge about other fish and even two movies with names associated with fish. We can see the hardship of the \ourapproach method: 1) \textbf{Insufficient knowledge}: The utilized MMKG itself lacks relevant information, impeding its ability to provide effective knowledge for multimodal reasoning. 2) \textbf{Ambiguity of knowledge}: Inherent ambiguities in the knowledge itself may lead to retrieve unrelated knowledge. In this example, ``BIG FISH'' does not refer to the fish but rather to a movie, introducing ambiguity.
\end{document}

%% file: table/table8.tex
\begin{table}[t]
    \centering
    
    \small
    \begin{center}
    \resizebox{0.475\textwidth}{!}{
    \renewcommand{\arraystretch}{1}

\begin{tabular}{lc}
\toprule
      \textbf{Settings} & \textbf{Accuracy (\%) on samples} \\
\midrule
    Visual\_FLAN-T5-11B	&86.59\tiny{(+0.00)} \\ 
    + KG	 & 90.37\tiny{(+3.78)} \\
    + MMKG 	&91.78\tiny{(+5.19)} \\
    + Alignment &92.32\tiny{(+5.73)} \\
\bottomrule
    \end{tabular}
    }
    \end{center}
    \caption{Ablation study on the samlpes.}
    \label{tab:parameter}
\label{tab:samples}
\end{table}

%% file: table/table9.tex
\begin{table}[t]
    \centering
    
    \small
    \begin{center}
    \resizebox{0.475\textwidth}{!}{
    \renewcommand{\arraystretch}{1}

\begin{tabular}{lc}
\toprule
      \textbf{Settings} & \textbf{Hits@1 on MARS} \\
\midrule
    Visual\_LLaMA-2 7B	&0.286\tiny{(+0.000)} \\ 
    + KG	 & 0.352\tiny{(+0.066)} \\
    + MMKG 	&0.381\tiny{(+0.095)} \\
    + Alignment &0.394\tiny{(+0.108)} \\
\bottomrule
    \end{tabular}
    }
    \end{center}
    \caption{Ablation study on MARS \textit{test} set.}
    \label{tab:parameter}
    \vspace{-2mm}
\label{tab:MARS}
\end{table}

%% file: table/table7.tex
\begin{table}[t]
    \centering
    \small
    \begin{center}
    \resizebox{0.475\textwidth}{!}{
    \renewcommand{\arraystretch}{1}

\begin{tabular}{ll}
\toprule
      \textbf{Settings} & \textbf{Hits@1 on MARS} \\
\midrule
    MR-MKG (with MarKG)	 & 0.405\tiny{(0.000)} \\
    MR-MKG (with MMKG)	&0.384\tiny{(-0.021)} \\
    Base (no MMKG) &0.286\tiny{(-0.119)} \\
\bottomrule
    \end{tabular}
    }
    \end{center}
    \caption{Impact of different MMKG on MARS.}
    \label{tab:parameter}
\label{tab:MMKGs}
\end{table}

%% file: figures/vg.tex
\begin{figure}[t]
    \centering
    \includegraphics[width=1\columnwidth]{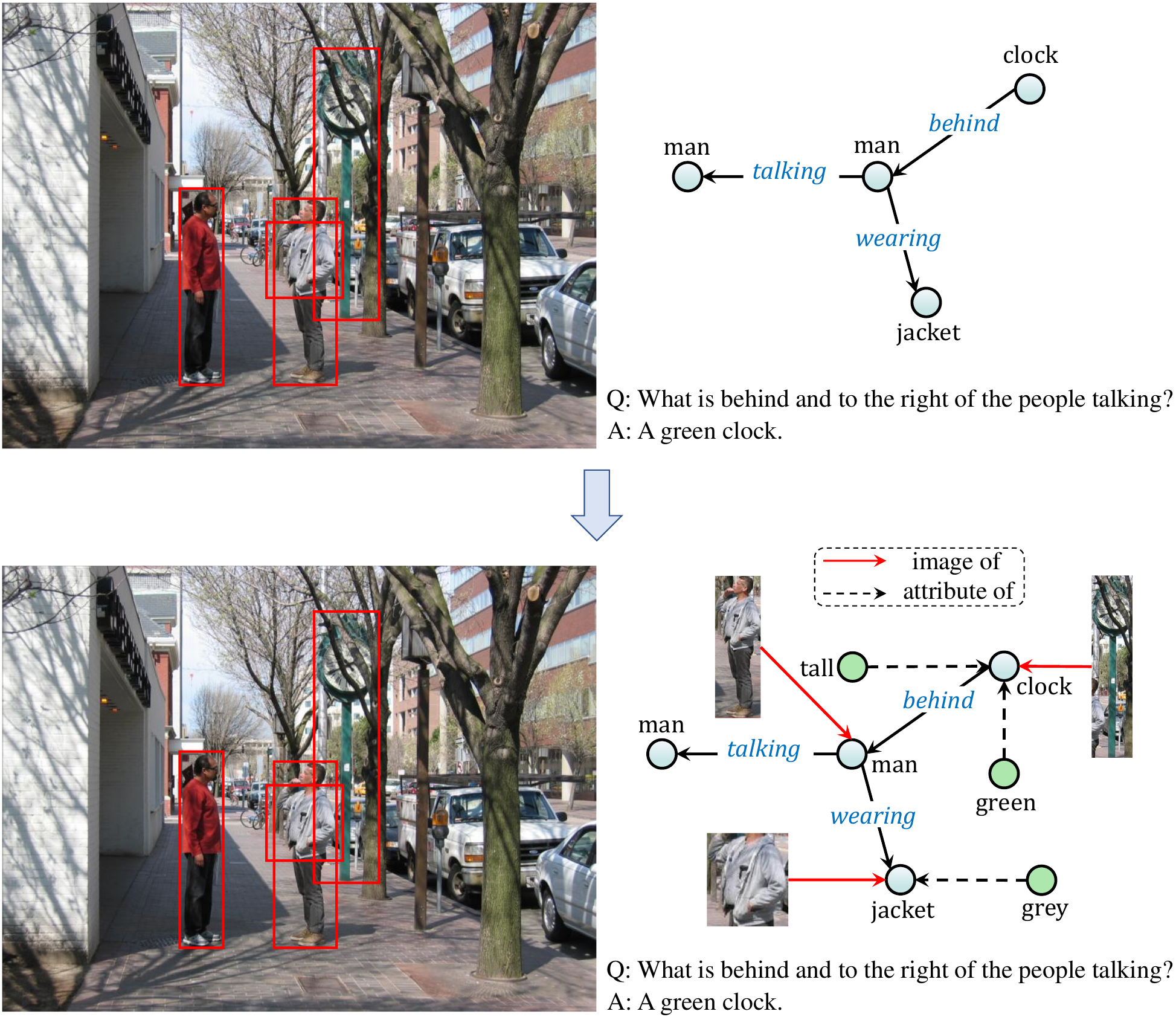}
    \caption{The process of transforming scene graph into a multimodal knowledge graph involves linking the object entities to their corresponding images and attributes.}
    \label{fig:vg}
\end{figure}

%% file: table/table6.tex
\begin{table}[t]
    \centering
    \small
    \begin{center}
    \resizebox{0.475\textwidth}{!}{
    \renewcommand{\arraystretch}{1}

\begin{tabular}{ccc}
\toprule
      Hyper-parameters & MarKG  & MARS \\
\midrule
    epoch & 3 & 3 \\
   \hspace{0.4cm}sequence length\hspace{0.4cm} & 96 &  128 \\
   learning rate & 2e-5 & 5e-6 \\
   batch size & 8 & 4 \\
   optimizer & AdamW & AdamW \\
   Weight decay & 0.01 & 0.01 \\
\bottomrule
    \end{tabular}
    }
    \end{center}
    \caption{Hyper-parameter settings of MARS training.}
    \label{tab:parameter}
\end{table}